\documentclass[letterpaper]{article} 
\usepackage{aaai2026}  
\usepackage{times}  
\usepackage{helvet}  
\usepackage{courier}  
\usepackage[hyphens]{url}  
\usepackage{graphicx} 
\usepackage{natbib}  
\usepackage{caption} 
\usepackage{multicol}
\usepackage{multirow}
\usepackage{booktabs}
\usepackage{xcolor}
\usepackage{colortbl}
\usepackage{algorithm}
\usepackage{algorithmic}

\usepackage{newfloat}
\usepackage{amsmath}
\usepackage{amssymb}
\usepackage{listings}

\DeclareCaptionStyle{ruled}{labelfont=normalfont,labelsep=colon,strut=off} 
\floatstyle{ruled}
\newfloat{listing}{tb}{lst}{}
\floatname{listing}{Listing}
\title{Tracing the Heart's Pathways: ECG Representation Learning from a Cardiac Conduction Perspective} 
\author{
    Tan Pan\textsuperscript{\rm 1, \rm 2}\thanks{\* Equal contribution. This work was conducted during an internship at the Shanghai Academy of Artificial Intelligence for Science.}, ~Yixuan Sun\textsuperscript{\rm 1, \rm 2}\footnotemark[1], Chen Jiang\textsuperscript{\rm 1, \rm 2}\thanks{\* Corresponding authors.}, Qiong Gao\textsuperscript{\rm 6}, Rui Sun\textsuperscript{\rm 1, \rm 2}, Xingmeng Zhang\textsuperscript{\rm 1, \rm 2},\\Zhenqi Yang\textsuperscript{\rm 2}, Limei Han\textsuperscript{\rm 1, \rm 2}, Yixiu Liang\textsuperscript{\rm 4, \rm 5, \rm 2}, Yuan Cheng\textsuperscript{\rm 1, \rm 2}\footnotemark[2], Kaiyu Guo\textsuperscript{\rm 2, \rm 3}\\
}
\affiliations{
    \textsuperscript{\rm 1} Fudan University \\
    \textsuperscript{\rm 2} Shanghai Academy of Artificial Intelligence for Science \\
    \textsuperscript{\rm 3} The University of Queensland \\
    \textsuperscript{\rm 4} Department of Cardiology, Zhongshan Hospital of Fudan University \\
    \textsuperscript{\rm 5} National Heart and Lung Institute, Imperial College London \\
    \textsuperscript{\rm 6} Department of Electrocardiology, Zhongshan Hospital of Fudan University

    \{pant23,sunyx23\}@m.fudan.edu.cn,~jiangchen@sais.org.cn,~cheng\_yuan@fudan.edu.cn
}

\begin{document}

\maketitle

\begin{abstract}
The multi-lead electrocardiogram (ECG) stands as a cornerstone of cardiac diagnosis. Recent strides in electrocardiogram self-supervised learning (eSSL) have brightened prospects for enhancing representation learning without relying on high-quality annotations. Yet earlier eSSL methods suffer a key limitation: they focus on consistent patterns across leads and beats, overlooking the inherent differences in heartbeats rooted in cardiac conduction processes, while subtle but significant variations carry unique physiological signatures. Moreover, representation learning for ECG analysis should align with ECG diagnostic guidelines, which progress from individual heartbeats to single leads and ultimately to lead combinations. This sequential logic, however, is often neglected when applying pre-trained models to downstream tasks. To address these gaps, we propose CLEAR-HUG, a two-stage framework designed to capture subtle variations in cardiac conduction across leads while adhering to ECG diagnostic guidelines. In the first stage, we introduce an eSSL model termed \textbf{C}onduction-\textbf{LEA}d \textbf{R}econstructor (CLEAR), which captures both specific variations and general commonalities across heartbeats. Treating each heartbeat as a distinct entity, CLEAR employs a simple yet effective sparse attention mechanism to reconstruct signals without interference from other heartbeats. In the second stage, we implement a \textbf{H}ierarchical lead-\textbf{U}nified \textbf{G}roup head (HUG) for disease diagnosis, mirroring clinical workflow. Experimental results across six tasks show a 6.84\% improvement, validating the effectiveness of CLEAR-HUG. This highlights its ability to enhance representations of cardiac conduction and align patterns with expert diagnostic guidelines.

\end{abstract}

\begin{links}
    \link{Code}{https://github.com/Ashespt/CLEAR-HUG}
\end{links}

\begin{figure}[!htbp]
  \centering
  \includegraphics[width=\columnwidth]{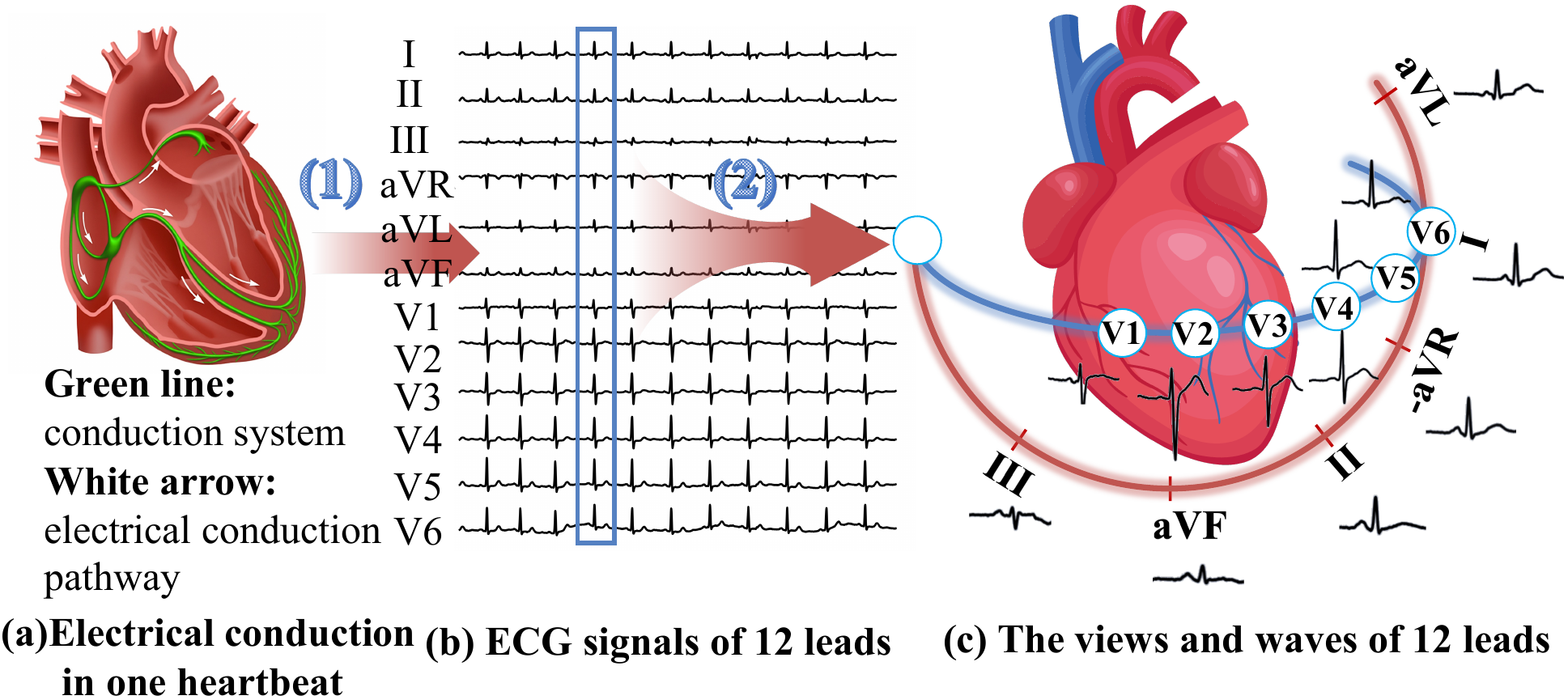}
  \caption{Cardiac conduction and the relationship between the 12 ECG leads: (1) Electrical activity propagates through the heart (a) and reflects on the 12 leads in the same time window (b). (2) The 12 leads capture the heart's electrical activity from different views (c).}
  \label{fig:figure1}
\end{figure}

\section{Introduction}
 ECG plays a crucial role in the diagnosis of various cardiac conditions~\cite{berkaya2018survey,auer2012association}. In recent years, AI-assisted ECG analysis has demonstrated significant potential in enhancing diagnostic accuracy~\cite{zhou2024open,poterucha2025detecting} and enabling real-time ECG monitoring~\cite{xia2013cloud,alimbayeva2024wearable}. To leverage large-scale unannotated data while addressing the dependency on expert knowledge and high-quality annotations, recent advances in ECG self-supervised learning (eSSL)~\cite{naguiding,jin2025reading} have enhanced ECG signal analysis.\par
 Self-supervised learning (SSL) methods for medical data can be divided into two approaches: contrastive learning~\cite{pan2025structure,wu2024voco} and generative learning~\cite{wang2023swinmm,hatamizadeh2021swin}.  Advanced eSSL methods build on these techniques by incorporating spatial and temporal priors to improve representation power~\cite{naguiding, zhang2022maefe}. For instance, ST-MEM uses a lead-wise shared decoder for better lead differentiation~\cite{naguiding}, while~\cite{zhang2023self} applies cross-reconstruction across time and frequency domains. These methods typically partition signals uniformly, whereas HeartLang~\cite{jin2025reading} partitions signals by heartbeats, treating them like words in language models. 
 


While existing methods demonstrate promising performance, they primarily focus on consistent patterns across leads and beats, which leads to the following two failure cases: \textbf{(1) Occasional anomalies.} For some abnormal cardiac conditions, the corresponding ECG signals lack consistently periodic or cyclic patterns. For instance, ventricular premature beats are characterized by abnormal contractions that arise prematurely within the cardiac cycle. These beats do not repeat consistently in every heartbeat and appear sporadically. \textbf{(2) Various focuses of leads across different diseases.} For example, premature ventricular contractions (PVCs) are characterized by premature, wide QRS complexes without a preceding P wave, typically seen in leads V1 and V2. Right bundle branch block (RBBB), on the other hand, shows an RSR' pattern in V1 and a broad R wave in V6, highlighting the focus on different lead combinations for diagnosis.\par
The aforementioned cases can be summarized as two problems: 1) the heartbeat- and lead-specific variations; 2) diagnosis-driven lead combination. To address these challenges, we reformulate the  ECG recording process from the perspective of cardiac conduction and propose a two-stage framework for learning representations progressively, first from heartbeats to leads at the pretraining stage and then from leads to their combinations at the finetuning stage. \par
During the pretraining stage, the relation between heartbeats and leads serves as auxiliary information. Specifically, as shown in Fig. \ref{fig:figure1} (1) and (2), this information primarily exhibits two characteristics: (1) the 12 leads corresponding to the same heartbeat share the same time window and the same electrical activity, which enables them to provide complementary views of the heart’s electrical activity to each other; (2) it can be observed that a single heartbeat consists of signals in two dimensions: heartbeat-specific and lead-specific, which leads to our assumption of conduction-guided and view-guided information. Based on the observation, we propose a reconstruction framework, termed \textbf{C}onduction-\textbf{LEA}d \textbf{R}econstructor (CLEAR), which aims to learn meaningful representations through recovering the signal of a heartbeat by utilizing both conduction-guided and view-guided information. Particularly, we propose a sparse attention mechanism with a tailored attention mask to highlight the conduction-guided and view-guided information through the reconstruction. \par
In the finetuning stage, the design of our method is guided by the clinical diagnosis workflow, enhancing the explainability of our model. As per the established ECG analysis guidelines~\cite{kligfield2007recommendations}, cardiac conditions are diagnosed at multiple levels: individual heartbeats, single-lead signals, and combinations of leads, which motivates us to design a multi-level linear model for specific diagnosis tasks. Specifically, we propose to integrate the representations learned from different lead combinations through various groups of linear models, which is termed as \textbf{H}ierarchical lead-\textbf{U}nified \textbf{G}roup (HUG) head. By simulating the clinical diagnosis, this design demonstrates more promising experimental results compared with conventional methods.\par

As mentioned above, each stage in our framework is motivated by a novel and reasonable insight. Therefore, the main contributions of this paper are threefold:\par
\textit{(1) Pretraining stage:} We introduce the concept of conduction-guided and view-guided information during pretraining, observed on the relationship between heartbeats and leads. Based on the insight, we propose a novel eSSL framework, CLEAR, to learn meaningful representations with conduction-guided and view-guided information.\par
\textit{(2) Finetuning stage:} We introduce a novel design inspired by the clinical diagnostic workflow to better align model finetuning with ECG analysis guidelines. Based on the design, we propose a linear model HUG to learn integral representations from different lead combinations.\par
\textit{(3) Experimental improvements:} Our method outperforms state-of-the-art (SOTA) techniques across 6 datasets, with an average improvement of 6.8\%. Notably, CLEAR-HUG achieved an 8.25\% improvement under 1\% of the training data, highlighting its strong potential for few-shot transfer. 
The code will be available upon publication.

\section{Related Work}
\textbf{Self-supervised Learning.} Current self-supervised learning (SSL) methods can be
classified into contrastive-based and reconstruction-based methods. For the contrastive-learning paradigm, SimCLR~\cite{chen2020simple} focuses on learning representations by maximizing agreement between augmented views of the same data instance and minimizing agreement between different instances. On the other hand, BYOL~\cite{grill2020bootstrap} enhances contrastive learning by eliminating the need for negative samples, leveraging a bootstrap mechanism. Similarly, SimSiam~\cite{chen2021exploring} avoids the use of negative pairs entirely and instead focuses on a simple predictor network to learn useful representations. In contrast, MoCo-v3~\cite{chen2021empirical} incorporates momentum encoders and dynamic dictionaries to further refine the learning of representations. While contrastive-based methods heavily rely on data augmentations, masked autoencoder (MAE)~\cite{he2022masked} follows a different approach, learning meaningful image representations by reconstructing missing or masked parts of the input data.\par
\textbf{Self-supervised Learning For ECG Signals.}
ESSL has made significant strides in recent years, with methods primarily falling into contrastive learning and generative approaches. CLOCS~\cite{kiyasseh2021clocs} utilizes multi-lead ECG temporal alignment to create positive pairs, while TS-TCC~\cite{eldele2021time} introduces temporal contrastive strategies, excelling in ECG classification tasks. Unlike contrastive learning, generative methods learn representations through reconstruction tasks. ST-MEM~\cite{naguiding} treats ECG as a spatiotemporal 2D signal for joint masking, while CRT~\cite{zhang2023self} performs cross-domain reconstruction in both time and frequency domains. Recently, HeartLang~\cite{jin2025reading} proposes that partitioning ECG signals following the heartbeat and learning it with heartbeat vocabulary. These approaches directly model data distributions and tend to learn global representations across all leads. In contrast, our work focuses on modeling the cardiac conduction, which carries unique physiological signatures.

\section{Insights from Cardiac Conduction}
\label{preliminary}

This section outlines ECG pattern insights and clinical guidelines for ECG diagnosis based on cardiac conduction features, which form our method's foundation. We define notations for a 12-lead ECG, though our approach is applicable to ECGs with any number of leads.



\noindent\textbf{Insight from ECG Patterns.} Based on the cyclic pattern of heartbeats and the principles of ECG signal acquisition~\cite{mirvis2001electrocardiography}, ECG signals exhibit inter-lead heartbeat conduction synchronization (Fig.~\ref{fig:figure1} (1)) and intra-lead heterogeneity (Fig.~\ref{fig:figure1} (2)). Specifically, during a single heartbeat, the ECG signals recorded by the 12 leads $\{L_i\}_{i=1}^{12}$ are derived from the same physiological process (one cardiac conduction), capturing the electrical activities of the heart at different views.  Within the $j$-th {heartbeat signal $b_i^j$} of the $i$-th lead, the formulations of the signals from the information of two perspectives are presented as follows:\par
\begin{figure*}[!t]
	\centering
	\includegraphics[width=0.9\linewidth]{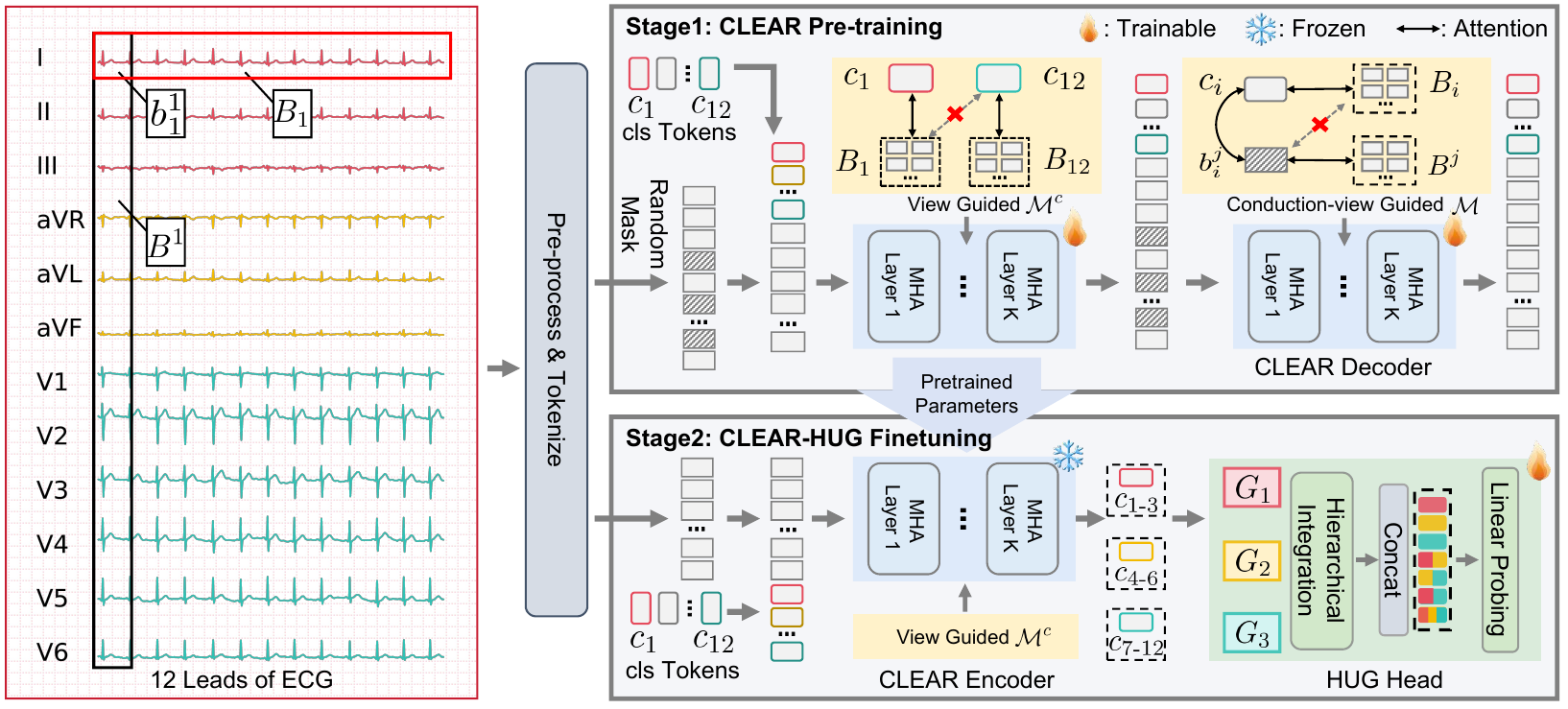}
	\caption{Illustration of proposed CLEAR-HUG framework. CLEAR-HUG is composed of two stages: (1) CLEAR Pre-training stage to learn specified representations of 12 leads, and (2) CLEAR-HUG Finetuning to integrate the lead feature from the pretrained encoder and simulate the clinical diagnosis procedure to provide predictions for downstream tasks. In which MHA layer stands for multi-head attention layer~\cite{vaswani2017attention}.}
	\label{fig:method}
    \vskip -0.1in
\end{figure*}
\textit{(1) Conduction-guided perspective}: The information $I_c$ originates from the same heartbeat conduction, reflecting the temporal synchronization shared across all leads during the $j$-th heartbeat. Thus, the set of signals across all leads within the $j$-th heartbeat can be denoted as $B^j = \{b_i^j\}_{i=1}^{12}$.\par
\textit{(2) View-guided perspective}: The information $I_v$ captures the spatial heterogeneity among leads, where each lead $L_i$ provides a distinct view of cardiac activity across all sampled heartbeats. The set of signals within the lead $L_i$ can be denoted as $B_i =\{b_i^j\}_{j=1}^{N}$, where $N$ represents the total number of sampled heartbeats.

\noindent\textbf{The lead combination guidelines of ECG diagnosis.} 
Considering ECG diagnosis tasks $T$, some tasks can be addressed using single-lead signals, while others require the integration of information across multiple leads. According to the recommendations for the standardization and interpretation of ECG~\cite{kligfield2007recommendations}, the 12 leads can be categorized into three groups: the first group $G_1 = \{I, II, III\}$ consists of bipolar limb leads; the second group $G_2 = \{\mathrm{aVR}, \mathrm{aVL}, \mathrm{aVF}\}$ comprises unipolar augmented limb leads; and the third group $G_3 = \{\mathrm{V1}, \mathrm{V2}, \mathrm{V3}, \mathrm{V4}, \mathrm{V5}, \mathrm{V6}\}$ includes the precordial leads. More details can be found in the Appendix. To emulate the diagnostic reasoning of clinicians, we propose a hierarchical and explainable model structure in the following section to enhance diagnostic performance during the fine-tuning stage.

\section{Method}
\label{sec:method}
Similar to previous SSL methods~\cite{tsai2020self, wang2022rethinking}, we divide representation learning into two stages: (1) pretraining to learn individual lead representations, and (2) fine-tuning to learn task-driven representations of lead combinations, simulating the diagnosis workflow.
\subsection{Preliminary}
\noindent\textbf{Masked Autoencoder.} We present the formulation of masked autoencoder (MAE)~\cite{he2022masked} tailored for sequence modeling, which constitutes a core component and the baseline of our approach. MAE is a widely used self-supervised learning framework that learns meaningful representations by reconstructing missing parts of the input. Following MAE, we formulate the framework as follows. Given an input sequence $\mathcal{B} = (b_1, b_2, \ldots, b_T)$ and a random masked position $k$, the modified sequence can be formulated as $\mathcal{B}_{masked} = (b_1, b_2, \ldots,\beta_k, \ldots, b_T)$, where $\beta_k$ denotes the masked element filled with zero values at the $k$-th position in the sequence. Note that there is more than one masked position within the input sequence in practice.  Given an encoder $\mathcal{E}$ and a decoder $\mathcal{D}$, the reconstructed data can be represented as $\hat{\mathcal{B}}=\mathcal{D}(\mathcal{E}(\mathcal{B}_{masked}))$. Then, the learning object of MAE is:
\begin{align}
   \mathcal{L}=\mu({||\hat{\mathcal{B}}-\mathcal{B}||_2^2}),
\end{align}
where $\mu(\cdot)$ is the average operator.\par
\noindent\textbf{Attention Mask.} 
Attention mask~\cite{vaswani2017attention} is a technique suitable for next-token prediction tasks to avoid data leak, which is widely used in attention layers of transformer models. Given an input feature $X$, the procedure of the self-attention mechanism can be formulated as follows:
\[
ATN(X)=softmax(\frac{XW_QW_K^\top X^\top}{\sqrt{d_k}})XW_V,
\]
where $W_Q, W_K,$ and $W_V$ are the weights for query, key, and value in the attention. $d_k$ is the dimension of the key weight. To capture the important information and accelerate the computation, the mask technique is applied in the attention, which can be formulated as follows:
\begin{equation}
ATN(X,\mathcal{M})=softmax(\frac{XW_QW_K^\top X^\top}{\sqrt{d_k}}+\mathcal{M})XW_V,
\label{eq:maskatn}
\end{equation}
where $\mathcal{M}$ is a $n_t\times n_t$ matrix, where $n_t$ is the number of the tokens in $X$. Specifically, $\mathcal{M}_{[i,j]}=-\infty$ represents ignoring position $(i,j)$ and $\mathcal{M}_{[i,j]}=0$ represents keep the position $(i,j)$ during attention computation.
\subsection{Pretraining: Conduction-Lead Reconstructor}
To capture individual heartbeats' details, we first segment a lead $L_i$ into a set of beats (tokens) $B_i=\{b_i^j\}_{j=1}^N$, where $i\in\{1,\ldots,12\}$. This process resembles tokenization, similar to the strategy in prior work~\cite{jin2025reading}.

Motivated by the conduction-view guided perspectives, we assume that a masked token can be recovered by conduction-guided information $I_c$ and view-guided information $I_v$. We define $(L_i,j)$ as $j$-th beat of $i$-th lead $L_i$ and the relationship can be formulated as $b_i^j=f(I_c^{(\cdot,j)},I_v^{(L_i,\cdot)})$, where $I_c^{(\cdot,j)}$ represents the conduction-guided information of the $j$-th beat and $I_v^{(L_i,\cdot)}$ represents the view-guided information of the $L_i$, $f$ is a mapping to reconstructed beat from corresponding information. Specifically, for the conduction-guided $I_c^{(\cdot,j)}$, we assume that $B^j$ can reflect the characteristics of $j$-th heartbeat conduction; for view-guided information $I_v^{(L_i,\cdot)}$, we introduce $cls$ tokens for 12 leads $C=\{c_i\}_{i\in\{1,2,...,12\}}$ to capture the global context of each view of cardiac activity. Thus, the reconstructed beat $\hat{b}_i^j$ can be:
\begin{equation}
\hat{b}_i^j=f(B^j,c_i).
\end{equation}
This formulation represents that a heartbeat token can be reconstructed by its same heartbeats but different viewed (lead) tokens and the specific global context of the lead. Therefore, the problem becomes how to learn the lead-specific representation $c_i$ and the synchronized heartbeat representation shared by $B^j$.

Based on the above discussion, we propose a \textbf{C}onduction-\textbf{LEA}d \textbf{R}econstructor (\textbf{CLEAR}) to build our pretraining framework, learning representations from heartbeats to leads by the relationship between $b_i^j$, $B^j$, and $c_i$. We first randomly initialize 12 $cls$ tokens $C=\{c_i\}_{i\in\{1,2,...,12\}}$. If we mask the token $b_i^j$, we hope it can be reconstructed only by $j$-th heartbeat information and the view-guided information $c_i$, like Equation 3. Thus, \textbf{The reconstruction process of $b_i^j$ should be restricted to interactions with the set ${B^j,c_i}$. For $c_i$, it should ``see" only the tokens $B_i$ from the same lead $L_i$, ensuring that it maintains both lead-specific and global representations.} To achieve this sparse relationship and interactions, CLEAR utilizes a sparse attention mechanism as shown in Stage 1 of Fig.~\ref{fig:method}\par
To be specific, the input $X\in\mathcal{R}^{12(N+1)\times d_t}$, where $d_t$ is the dimension of the tokens, consists of the tokens from 12 leads and the initialized 12 \textit{cls} tokens. Specifically, $X=[C,B_1,B_2,\ldots,B_{12}]$. Given the input $X$ and the masked tokens $\{\beta_k, k\in K\}$, where $K\in[1,12N]$ is the position set of the masked tokens, we acquire masked input as $X_{masked}$. Then, we propose a new sparse attention mechanism tailored for the ECG data in the transformer. Specifically, we separate the attention mask $\mathcal{M} = [m_{p,q}]_{(12N+12)\times(12N+12)}$ into two parts: $\mathcal{M}^c = [m_{p,q}]_{12\times(12N+12)}$ and $\mathcal{M}^b = [m_{p,q}]_{12N\times(12N+12)}$.  The attention mask for \textit{cls} tokens $\mathcal{M}^c$ is defined as follows:
\begin{align} 
    \mathcal{M}^c_{[p,q]}\!\!\triangleq \!\!\left\{ 
    \begin{aligned}
    &0, &if~q\in Q_p\\&&or~p=q\\
    &-\infty,&else.
    \end{aligned}
    \right.
\end{align}
 where $Q_p = [(p-1)N+13,pN+12], Q_p \subset Z^+$. And, the attention mask for heartbeat tokens $\mathcal{M}^b$ is defined as:
\begin{align}  \mathcal{M}^b_{[p,q]}\!\!\triangleq \!\!\left\{ 
    \begin{aligned}
    &0, &if~p\in K, q\in R_p\\&&or~p\notin K, q\in C_p\\
    &&or~q=\left\lfloor \frac{p-1}{N} \right\rfloor+1\\
    &-\infty,&else.
    \end{aligned} 
    \right.
\end{align}
where $R_p = \{o+12| o\bmod12=p\bmod12,o\leq12N\, o\in Z^+\}$, i.e., the position indicates the same heartbeat, $C_p=\{o+12|o\bmod N=o\bmod N, o\leq12N, o\in Z^+\}$, i.e., the position indicates the same lead. 
The final attention mask $\mathcal{M}$ is defined as $\begin{bmatrix}
\mathcal{M}^c \\ \mathcal{M}^b
\end{bmatrix}$.\par
The view-guided and conduction-guided attention focuses on specific tokens as per the equation. $\mathcal{M}^c$ is applied to both $\mathcal{E}$ and $\mathcal{D}$, while $\mathcal{M}$ is applied only to $\mathcal{D}$, as $\mathcal{E}$'s input doesn't contain masked tokens. Finally, the reconstruction loss of the masked input $X_{masked}$ can be formulated as:
\begin{align}
    \mathcal{L}=\|\mathcal{D}(\mathcal{E}(X_{masked}))- X \|_2^2,
\end{align}
\subsection{Finetuning: Hierarchical Lead-unified Group}
According to the insights of lead combination guidelines in ECG diagnosis discussed above, we propose a lead grouping strategy, \textbf{H}ierarchical lead-\textbf{U}nified \textbf{G}roup (\textbf{HUG}), as shown in Fig.~\ref{fig:hug}, to adapt the pretrained eSSL model CLEAR to downstream tasks. As shown in Fig.~\ref{fig:figure1}, HUG has 7 linear heads $\{\phi_i\}_{i=1}^7$ to simply and effectively learn information of the lead combinations, which aims to mimic the workflow of clinical diagnosis. Given 12 \textit{cls} tokens $\{c_i\}_{i\in\{1,2,...,12\}}$ from the output of CLEAR Encoder, the finetuning pipeline can be summarized as 3 steps:
\begin{figure}[!h]
	\centering
	\includegraphics[width=0.85\linewidth]{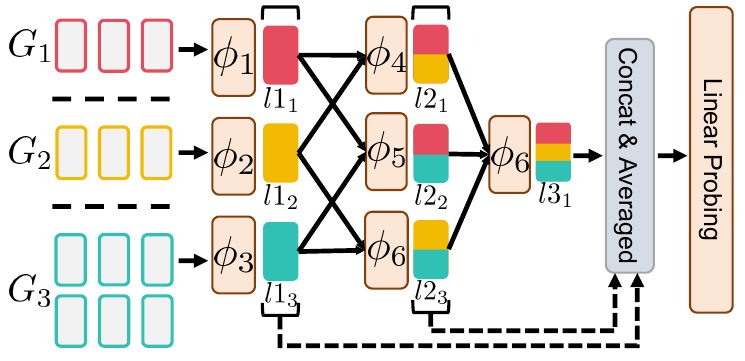}
	\caption{{Illustration of proposed HUG head.} The HUG head integrates the three ECG lead groups via a three-stage hierarchical framework.}
	\label{fig:hug}
    
\end{figure}

\noindent \textbf{Step 1:} Following principles of ECG signal acquisition~\cite{kligfield2007recommendations}, we divide \textit{cls} tokens of 12 leads into 3 groups by the guidance aforementioned, i.e., $G_1=\{I, II, III\}$, $G_2=\{aVR, aVL, aVF\}$, and $G3=\{V1, V2, V3, V4, V5, V6\}$. Each group is processed by a separate linear layer $\phi_i$, and the outputs are averaged:
\begin{align}
l1=\{\mu(\phi_1(G_1)),\mu(\phi_2(G_2)),\mu(\phi_3(G_3))\} \notag
\end{align}
\noindent \textbf{Step 2:} The outputs of Step 1 are reallocated in pairwise combinations. The recombined token pairs are denoted as $\mathcal{P} = \{(l1_1, l1_2), (l1_1, l1_3), (l1_2, l1_3)\}$, and each pair is processed by a separate head and then averaged:
\begin{align}
l2=&\{\mu(\phi_4(\mathcal{P}_1)),\mu(\phi_5(\mathcal{P}_2)),\mu(\phi_6(\mathcal{P}_3))\}, \notag
\end{align}

\noindent \textbf{Step 3:} Aggregate the outputs from Step 2:
\begin{align}
l3=\mu(\phi_7(l 2)). \notag
\end{align}


At last, the outputs of the 3 steps are concatenated to form the input of the final linear model. 
\begin{align}
f_g=\mu(l1\cup l2 \cup \{l3\}). \notag
\end{align}
The lead combination of the diagnostic process is emulated through this hierarchical group head, where the seven groups correspond to distinct lead combinations. The hierarchical structure enables progressive integration of information, with each level building upon the representations learned at the previous level.
\section{Experimental Setup}
We organized experiments according to the configuration in MERL~\cite{liu2024zero}. Following MERL's configuration, the CLEAR-HUG is first pre-trained on MIMIC-IV-ECG~\cite{gow2023mimic} and evaluated on downstream tasks. And we select the SSL methods, SimCLR~\cite{chen2020simple}, BYOL~\cite{grill2020bootstrap}, MoCo-v3~\cite{chen2021empirical}, and SimSiam~\cite{chen2021exploring}, time-series SSL method, TS-TCC~\cite{eldele2021time}, and eSSL methods, CLOCS~\cite{kiyasseh2021clocs}, ASTCL~\cite{wang2023adversarial}, CRT~\cite{zhang2023self}, ST-MEM~\cite{na2024guiding}, and HeartLang~\cite{jin2025reading} for comparison.
\subsection{Setup of CLEAR Pre-training}
\textbf{Datasets.} Following previous works~\cite{zhang2023self, naguiding, jin2025reading}, we use MIMIC-IV-ECG~\cite{gow2023mimic} to pre-train CLEAR-HUG. This dataset comprises 161,352 subjects with 800,035 12-lead ECG recordings. Each sample lasts for 10 seconds at 500 Hz. We further process the dataset with (1) splitting the signal according to heartbeat detection; (2) replacing the “NaN” and “Inf” values in the ECG recordings with the average of six neighboring points, similar to HeartLang~\cite{jin2025reading}.

\noindent\textbf{Implementation Detail.} 
We downsample all the ECG records of MIMIC to 100 Hz and tokenize each lead as 15 heartbeat tokens with a $[cls]$ token using the designed tokenizer. We set the learning rate to $5\times10^{-4}$ and trained for 100 epochs. We set the mask ratio of input tokens as 80\%. Analysis of the hyperparameter can be found in the appendix. The AdamW optimizer and cosine annealing scheduler are applied for learning rate adjusting. All experiments were conducted on 6 NVIDIA A100 GPUs, and the batch size is set to 256 for each GPU.  




\begin{table*}[!t]
\centering
\setlength\tabcolsep{0.0052\linewidth}
\renewcommand\arraystretch{1}
 \scalebox{0.7}{\begin{tabular}{ccrrrrrrrrrrrrrrrrrr}
\toprule
\multirow{2}{*}{\textbf{Method}} & \multirow{2}{*}{\textbf{Source}} & \multicolumn{3}{c}{\textbf{PTBXL-Sub}} & \multicolumn{3}{c}{\textbf{PTBXL-Super}} & \multicolumn{3}{c}{\textbf{PTBXL-Form}} & \multicolumn{3}{c}{\textbf{PTBXL-Rhythm}} & \multicolumn{3}{c}{\textbf{CPSC2018}} & \multicolumn{3}{c}{\textbf{CSN}} \\
                        & & \multicolumn{1}{c}{1\%}       & \multicolumn{1}{c}{10\%}     & \multicolumn{1}{c}{100\%}    & \multicolumn{1}{c}{1\%}       & \multicolumn{1}{c}{10\%}     & \multicolumn{1}{c}{100\%}  &  \multicolumn{1}{c}{1\%}       & \multicolumn{1}{c}{10\%}     & \multicolumn{1}{c}{100\%}    &  \multicolumn{1}{c}{1\%}       & \multicolumn{1}{c}{10\%}     & \multicolumn{1}{c}{100\%}   &  \multicolumn{1}{c}{1\%}       & \multicolumn{1}{c}{10\%}     & \multicolumn{1}{c}{100\%}  &  \multicolumn{1}{c}{1\%}       & \multicolumn{1}{c}{10\%}     & \multicolumn{1}{c}{100\%} \\ \midrule
SimCLR & ICML2020 & 60.84 & 68.27 & 73.39 & 63.41 & 69.77 & 73.53 & 54.98 & 56.97 & 62.52 & 51.41 & 69.44 & 77.73 & 59.78 & 68.52 & 76.54 & 59.02 & 67.26 & 73.20 \\

BYOL & NeurIPS2020 & 57.16 & 67.44 & 71.64 & 71.70 & 73.83 & 76.45 & 48.73 & 61.63 & 70.82 & 41.99 & 74.40 & 77.17 & 60.88 & 74.42 & 78.75 & 54.20 & 71.92 & 74.69 \\


MoCo-v3 & ICCV2021 & 55.88 & 69.21 & 76.69 & 73.19 & 76.65 & 78.26 & 50.32 & 63.71 & 71.31 & 51.38 & 71.66 & 74.33 & \underline{62.13} & 76.74 & 75.29 & 54.61 & 74.26 & 77.68 \\

SimSiam & CVPR2021 & 62.52 & 69.31 & 76.38 & 73.15 & 72.70 & 75.63 & 55.16 & 62.91 & 71.31 & 49.30 & 69.47 & 75.92 & 58.35 & 72.89 & 75.31 & 58.25 & 68.61 & 77.41 \\

TS-TCC & IJCAI2021 & 53.54 & 66.98 & 77.87 & 70.73 & 75.88 & 78.91 & 48.04 & 61.79 & 71.18 & 43.34 & 69.48 & 78.23 & 57.07 & 73.62 & 78.72 & 55.26 & 68.48 & 76.79 \\


CLOCS & ICML2021 & 57.94 & 72.55 & 76.24 & 68.94 & 73.36 & 76.31 & 51.97 & 57.96 & 72.65 & 47.19 & 71.88 & 76.31 & 59.59 & \underline{77.78} & 77.49 & 54.38 & 71.93 & 76.13 \\ 

ASTCL & TNNLS2024 & 61.86 & 68.77 & 76.51 & 72.51 & 77.31 & 81.02 & 44.14 & 60.93 & 66.99 & 52.38 & 71.98 & 76.05 & 57.90 & 77.01 & 79.51 & 56.40 & 70.87 & 75.79 \\

CRT & TNNLS2023 & 61.98 & 70.82 & 78.67 & 69.68 & 78.24 & 77.24 & 46.41 & 59.49 & 68.73 & 47.44 & 73.52 & 74.41 & 58.01 & 76.43 & \underline{82.03} & 56.21 & \underline{73.70} & 78.80 \\

ST-MEM & ICLR2024 & 54.12 & 57.86 & 63.59 & 61.12 & 66.87 & 71.36 & 55.71 & 59.99 & 66.07 & 51.12 & 65.44 & 74.85 & 56.69 & 63.32 & 70.39 & \underline{59.77} & 66.87 & 71.36 \\
Heartlang & ICLR2025 & \underline{64.68} & \underline{79.34} & \underline{88.91} & \underline{78.94} & \underline{85.59} & \underline{87.52} & \underline{58.70} & \underline{63.99} & \underline{80.23} & \underline{62.08} & \underline{76.22} & \underline{90.34} & 60.44 & 66.26 & 77.87 & 57.94 & 68.93 & \underline{82.49} \\ \midrule \midrule
{CLEAR} & {Ours} &  73.86 & 80.11 & 89.3 & 78.59 & 85.43 & 88.68 & 61.00 & 69.96 & 80.34 & 79.24 & 86.72 & 93.66   &  64.82  &  78.10   &  89.59   &  62.88  &  76.33 &  86.75  \\ 
\textbf{Gains(\%)}   & - & \textbf{+9.18}      & \textbf{+0.77}     & \textbf{+0.39}     & -0.35    & -0.16     & \textbf{+1.16}     & \textbf{+2.30}      & \textbf{+5.97}      & \textbf{+0.11}     & \textbf{+17.16}     & \textbf{+10.5}     & \textbf{+3.32}    &     \textbf{+2.69}     &    \textbf{+0.32}      &     \textbf{+7.56}     &    \textbf{+3.11}   &    \textbf{+2.63}    &    \textbf{+4.26}
\\
\rowcolor[HTML]{EFEFEF}
CLEAR-HUG   & {Ours} & 76.66     & 84.59    & 91.44    & 79.61    & 86.85    & 90.24    & 61.01     & 75.19     & 82.86    & 79.48    & 90.57    & {94.08}   &     66.97    &    82.59     &     91.17     &    72.09   &    82.14    &    89.93\\
\rowcolor[HTML]{EFEFEF}
\textbf{Gains(\%)}  & - & \textbf{+11.98}      & \textbf{+5.25}     & \textbf{+2.53}     & \textbf{+0.67}    & \textbf{+1.26}     & \textbf{+2.72}     & \textbf{+2.31}      & \textbf{+11.2}      & \textbf{+2.63}     & \textbf{+17.4}     & \textbf{+14.35}     & \textbf{+3.74}    &     \textbf{+4.84}     &    \textbf{+4.81}      &     \textbf{+9.14}     &    \textbf{+12.32}   &    \textbf{+8.44}    &    \textbf{+7.44}
\\   \bottomrule
\end{tabular}}
\caption{{Comparison of proposed framework with other eSSL methods on six downstream tasks.} We designed two settings of our proposed framework, the first is the pretrained CLEAR, followed by linear probing. The second is the fully CLEAR-HUG framework for linear probing.  The SOTA results are \underline{underlined} and gains for both settings are compared with them.}
\label{tab:res1}

\end{table*}
\begin{table*}[!t]
\centering
\setlength\tabcolsep{0.019\linewidth}
\renewcommand\arraystretch{1}
 \scalebox{0.7}{\begin{tabular}{cccccccccccccc}
\toprule
\multicolumn{2}{c}{\textbf{Method}} & \multicolumn{3}{c}{\textbf{PTBXL-Sub}} & \multicolumn{3}{c}{\textbf{PTBXL-Super}} & \multicolumn{3}{c}{\textbf{PTBXL-Form}} & \multicolumn{3}{c}{\textbf{PTBXL-Rhythm}} \\
Pre-training & Downstream & 1\%       & 10\%     & 100\%    & 1\%      & 10\%     & 100\%    & 1\%       & 10\%      & 100\%    & 1\%      & 10\%     & 100\%  \\ \midrule
w/o pretraining & N/A & 65.16 & 77.10 & 85.89 & 76.64 & 83.77 & 87.16 & 54.43 & 63.98 & 73.68 & 51.51 & 77.80 & 86.65 \\ 
Baseline & N/A & 67.49 & 77.68 & 89.00 & 74.49 & 84.07 & 87.33 & 55.68 & 64.13 & 77.06 & 73.73 & 83.19 & 91.52 \\ 
Baseline & HUG & 75.41  & 83.76  & 90.74  & 78.70  & 85.57  & 88.76  & 56.08  & 70.38  & 78.39  & 73.76  & 86.68  & 91.60 \\ 
CLEAR & N/A &  73.86 & 80.11 & 89.30 & 78.59 & 85.43 & 88.68 & 61.00 & 69.96 & 80.34 & 79.24 & 86.72 & 93.66  \\\midrule
\rowcolor[HTML]{EFEFEF}
CLEAR & HUG  & \textbf{76.66}     & \textbf{84.59}    & \textbf{91.44}    & \textbf{79.61}    & \textbf{86.85}    & \textbf{90.24}    & \textbf{61.01}     & \textbf{75.19}     & \textbf{82.86}   & \textbf{79.48}    & \textbf{90.57}    & {\textbf{94.08}} \\   \bottomrule
\end{tabular}}
\caption{{Results of the ablation study of modules for CLEAR-HUG}. Best scores are in bold. The pre-training baseline is a masked autoencoder.
`w/o pretraining' denotes direct finetuning from scratch with the downstream head; SLG denotes lead grouping without hierarchy. `N/A' (downstream) denotes linear probing on the frozen pre-trained backbone.}
\label{tab:res2}
\end{table*}

\subsection{Setup of Downstream Funetuning}
\textbf{Datasets.} We conduct experiments on downstream tasks using three publicly accessible datasets: PTB-XL~\cite{wagner2020ptb}, CPSC2018~\cite{liu2018open}, and Chapman-Shaoxing-Ningbo (CSN)~\cite{zheng202012}.\par
\textit{PTB-XL}: PTB-XL includes 21,837 12-lead ECG recordings from 18,885 patients, each lasting 10 seconds with 500 Hz sampling rate. The dataset is divided into four subsets based on the SCP-ECG protocol: Form (19 classes), Rhythm (12 classes), Superclass (5 classes), and Subclass (23 classes). We followed the official configuration for splitting the training, validation, and test sets.\par
\textit{CPSC2018}: CPSC2018 contains 6,877 12-lead ECG recordings a 500 Hz sampling rate. Unlike PTB-XL, the ECG duration ranges from 6 to 60 seconds, with 9 labels. We split the data into training(70\%), validation (10\%), and testing (20\%) sets, following the HeartLang configuration. \par
\textit{CSN}: CSN includes 45,152 12-lead ECG recordings sampled at 500 Hz, each lasting 10 seconds. We use 23,026 recordings with 38 labels, excluding ``unknown" annotations, following MERL. The dataset is split into 70\% training, 10\% validation, and 20\% testing.

\noindent\textbf{Implementation Detail.} We down-sample the ECG signals to 100 Hz before fine-tuning CLEAR-HUG, using a pre-trained tokenizer to generate 16 tokens per lead. We set the HUG head, and the linear classifier is trainable. The HUG head and linear classifier are trainable, with the CLEAR Encoder initialized from pre-trained parameters. To compare with previous methods under low-resource conditions, we follow work~\cite{jin2025reading} using 1\%, 10\%, and 100\% of the training data for each task to finetune the modules.  The learning rate is set as $5\times10^{-3}$, and the training converged within 100 epochs. Besides, we scaled the ECG recordings to $[-3, 3]$ for CPSC2018 and CSN datasets, obtained all test results using the trained parameters with the best validation results, and used the macro AUC metric for evaluation, to maintain a fair comparison with previous works~\cite{zhang2023self, naguiding, jin2025reading}. 
\section{Results and Analysis}
\subsection{Primary Results}
Table~\ref{tab:res1} shows the comparison of our CLEAR-HUG with the SOTA methods. To present a fair comparison, we designed two settings, where the first directly uses the linear probing layer for downstream tasks using features from the pre-trained CLEAR Encoder. While the second introduced our proposed HUG head between the pretrained backbone and the linear probing layer to integrate features of 12 leads further. Compared with the SOTA methods, both CLEAR and CLEAR-HUG demonstrated significant advantages in almost all downstream tasks, with 3.94\% and 6.84\% improvements in the average separately. Notably, in the PTBXL-Rhythm task, both CLEAR and CLEAR-HUG outperform the SOTA results by over 17\% and 10\%, respectively, when fine-tuned on 1\% and 10\% of the training data. This result further confirms the advantages of our design for heartbeat conduction in representing and analyzing heartbeat rhythms.

Further comparing the CLEAR and CLEAR-HUG, we found that the designed lead fusion strategy can better overcome the challenges of more diverse category classification in ECG understanding tasks. For example, in the CSN dataset, when introducing our designed HUG, our CLEAR-HUG gains an additional performance improvement of over 9.21\%, 5.81\%, and 3.18\% when finetuning on 1\%, 10\%, and 100\% of the training data, compared to CLEAR. Furthermore, the performance of CLEAR-HUG is improved in all downstream task settings by a clear margin. We attribute this result to the fact that the hierarchical grouping strategy is more in line with the clinical diagnosis process and can better aggregate lead features to achieve adaptation to more complex scenarios and more diverse categories.

\subsection{Ablation Study}
\textbf{Designed Modules} To explore the potential necessity of the CLEAR pre-training and the HUG finetuning, we designed several ablation studies as shown in Table~\ref{tab:res2}. In the experiment, we considered four combinations of with/without CLEAR pre-training and with/without HUG head. Notably, our baseline is the native masked autoencoder model, having the same learnable parameters as CLEAR, but using full attention instead of the designed sparse attention.


From the perspective of CLEAR pretraining, experiments show our designed conduction-view guidance can effectively improve model performance, independent of the impact of the HUG head. We attribute this improvement to CLEAR pretraining's ability to better integrate both common and lead-specific information within each lead. This is crucial for tasks involving the identification of heartbeat patterns and rhythm regularity, such as heart rhythm analysis, where we observed a 17.4\% improvement with 1\% of the training data.
And for the Hierarchical Unified Group (HUG) head, its introduction improves model performance overall, with more pronounced benefits in fine-grained tasks such as sub-class categorization. We attribute this improvement to the grouping strategy, whose hierarchical structure better aligns with clinical diagnosis processes while enabling more comprehensive utilization of 12-lead information. We will further analyze the internal design of our model and provide a series of in-depth analyses in the following.

\begin{table}[!h]
\centering
\scalebox{0.7}{\begin{tabular}{lcll}
\toprule
\multirow{2}{*}{\textbf{Method}} & \multicolumn{3}{c}{\textbf{CSN}}                                    \\
                        & 1\% & \multicolumn{1}{c}{10\%} & \multicolumn{1}{c}{100\%} \\ \midrule
N/A (Averaged)         & 62.88 &  76.33   &    86.75   \\
Weighted Averaged        &  66.73   &    80.02                      &         88.11                  \\
Single-level Grouping                &   68.70  &   80.21                       &          89.35                 \\ \midrule
HUG Head                &   \textbf{72.09}  &   \textbf{82.14}                       &         \textbf{89.93}                  \\ \bottomrule
\end{tabular}}
\caption{Comparison of different designs to integrate 12 leads for downstream tasks.}
\label{tab:ablher}
\end{table}

\subsection{Insights of CLEAR-HUG}
\begin{figure}[]
  \centering
  \includegraphics[width=\columnwidth]{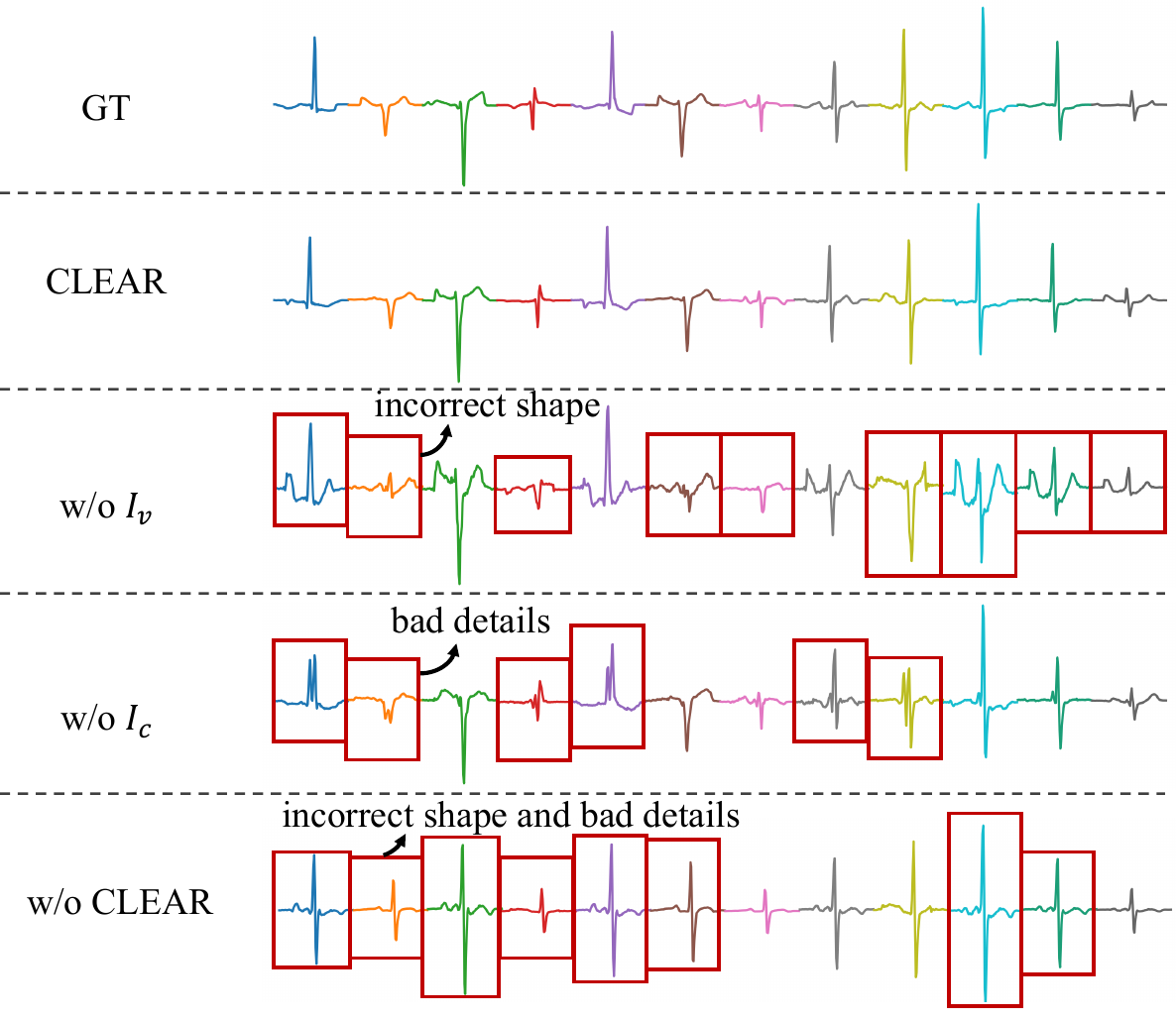}
  \caption{Visualization of reconstructed tokens for CLEAR variants w/o components: 12 leads in different colors, with poor reconstructions marked by red boxes.}
  \label{fig:Ic_Iv}
\end{figure}

\textbf{$I_v$ and $I_c$ visualization of pretrained model.} We visualize heartbeat reconstruction under four settings: CLEAR, w/o $I_v$, w/o $I_c$, and w/o CLEAR. As shown in Fig.~\ref{fig:Ic_Iv}, CLEAR yields the best reconstruction. We further analyze the reconstruction effect as follows: (1) When removing $I_v$ (only conduction-guided information), the reconstructed signal shows an incorrect wave shape, consistent with $I_v$ encoding the global lead context and overall waveform contour. (2) When removing $I_c$ (only view-guided information), the global shape is preserved but fine details are lost, supporting that $I_c$ captures heartbeat-specific details. (3) When both $I_c$ and $I_v$ are removed, neither shape nor details are correct, further validating our design of the two guidance terms.

\begin{figure}[!t]
  \centering
  \includegraphics[width=\columnwidth]{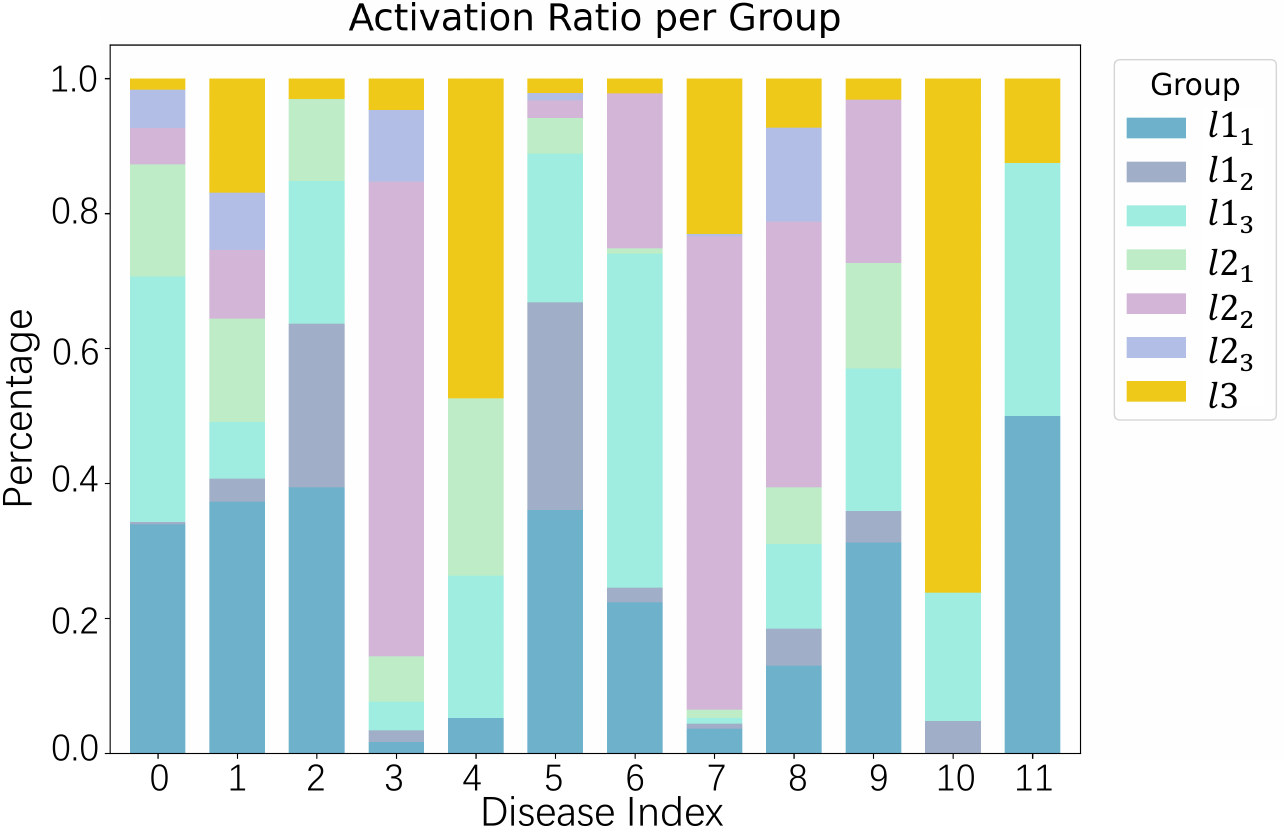}
  \caption{The visualization of activation ratios on 7 group combinations from HUG on different diseases.}
  \label{fig:group}
\end{figure}

\noindent\textbf{Grouping in finetuning.} To further investigate the impact of combining the three groups in HUG, we visualize the activation patterns of HUG on PTBXL-form in Fig. \ref{fig:group}. The visualization demonstrates that HUG exhibits diagnosis-specific activation patterns across its different group combinations, varying with distinct types of cardiac abnormalities. This differential activation directly corresponds to the feature of different cardiac conditions, confirming the ability of ClEAR-HUG to autonomously extract diagnosis-critical features. Further analysis can be found in the appendix.


\noindent\textbf{Hierarchical lead integration.} We also evaluate different integration strategies for grouping 12 leads. We select the CSN dataset as our benchmark and compare our HUG design with the average, weighted average, and single-level grouping methods. We note that single-level grouping integrates 3 groups through full combination and generates 7 combinations as HUG. Further ablation studies (Table \ref{tab:ablher}) demonstrate that the hierarchical lead-grouping (HUG) design outperforms alternative approaches by a clear margin. We attribute this to the fact that the hierarchical design is closer to clinical diagnosis and can more effectively utilize the information of the 12 leads. Results of other datasets are available in the appendix.

\section{Conclusion and Limitation}

In this paper, we propose CLEAR-HUG, which comprises: (1) CLEAR (Conduction-LEAd Reconstructor), which uses conduction-inspired sparse attention to guide masked 12-lead ECG token reconstruction; (2) HUG (Hierarchical lead-Unified Group head), which employs a hierarchical grouping strategy to mimic clinical diagnosis. Experiments on 6 datasets show that CLEAR-HUG surpasses existing methods across diverse ECG analysis tasks, and ablations verify the contribution of each component. \textbf{Limitations.} CLEAR-HUG could be extended to more downstream tasks. Moreover, HUG is tightly coupled with our pretraining design, making it incompatible with prior ECG foundation models that use only a single \textit{cls} token.

\section*{Acknowledgments}
This work was supported by Fudan University AI4S Project (FudanX24AI056), Pujiang Talent Program (24PJD014), and Shanghai Municipal Science and Technology Major Project (2023SHZDZX02 and 2017SHZDZX01 to L.J.). The computations in this research were performed using the CFFF platform of Fudan University.

\bibliography{aaai2026}

\newpage
\appendix
\section{Appendix}
\subsection{A. Grouping in finetuning}
\label{grouping}
\begin{figure}[]
	\centering
	\includegraphics[width=1.0\linewidth]{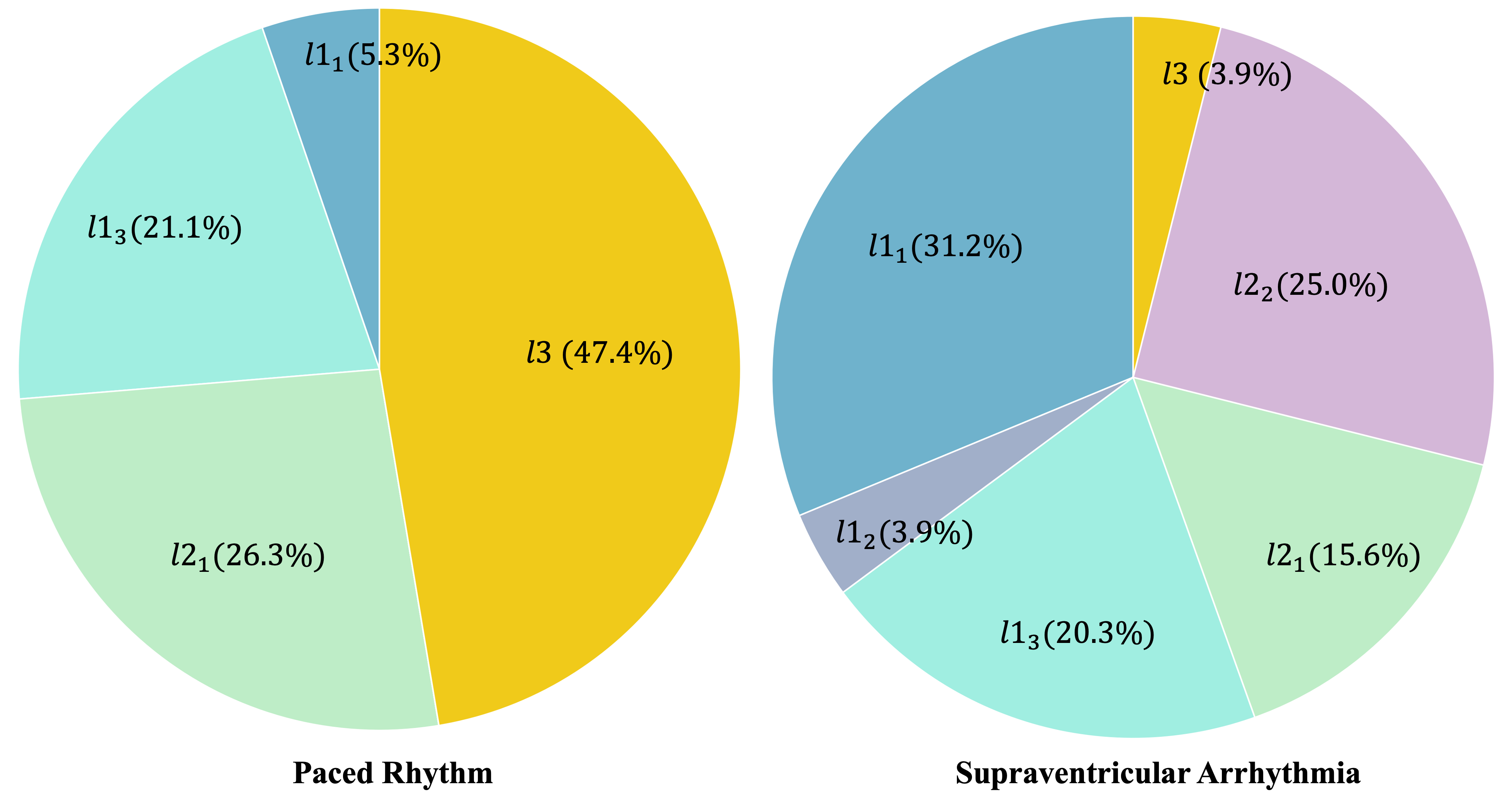}
    \caption{The pie charts of activation ratios across 7 group combinations from HUG for different diseases.}
    \label{fig:grouppie}
\end{figure}

In this section, we further discuss the result of the activation ratio of grouping in finetuning as shown in Fig.~\ref{fig:grouppie}. We discuss the disease correlated leads~\cite{kligfield2007recommendations} and corresponding lead combination, meanwhile, compared with our activation groups. 

For example, paced rhythm is diagnosed primarily by examining the following ECG leads:

\begin{itemize}
    \item $l1_1$ (I, II, III): Used to observe pacing-induced changes, particularly in lead II, where pacing spikes and abnormal QRS complexes are often seen.
    \item $l1_2$ (aVR, aVL, aVF): These leads can provide additional information, especially in aVR, where pacing effects may be more prominent.
    \item $l1_3$ (V1-V6): V1 and V2 help identify pacing-induced conduction patterns, while V3-V6 provide insight into QRS complex morphology.
    \item $l2_1$ ($l1_1$+$l1_2$): Combined leads offer a comprehensive view of pacing effects on the heart.
    \item $l3$ A full 12-lead is required to confirm the pacing site.
\end{itemize}

For another example, supraventricular arrhythmia is typically diagnosed by examining the following ECG leads:

\begin{itemize}
\item $l1_1$ (I, II, III): These leads are crucial for detecting irregular rhythms, particularly in lead II, where abnormalities like premature atrial contractions (PACs) or atrial tachycardia may be visible.

\item $l1_2$ (aVR, aVL, aVF): aVR is useful for assessing P-wave morphology and identifying atrial arrhythmias.

\item $l1_3$ (V1-V6): V1 is often used to identify atrial and junctional arrhythmias, while V2-V6 can help assess the spread of the arrhythmia across the ventricles.

\item $l2_1$ ($l1_1$+$l1_2$): Combined leads offer a broader perspective, helping to pinpoint the origin and propagation of supraventricular arrhythmias.
\end{itemize}

The above discussion aligns closely with our visualization of activation ratios across 7 groups in downstream tasks, further validating the clinical relevance of our HUG design. Tab.~\ref{tab:ecg_leads} shows the standard 12-Lead ECG electrode configuration~\cite{mirvis2001electrocardiography}.

\begin{table}[!t]
\centering
\setlength\tabcolsep{0.01\linewidth}
\scalebox{0.85}{\begin{tabular}{lll}
\toprule
\textbf{Lead Type} & \textbf{Positive Input} & \textbf{Negative Input} \\
\midrule
\multicolumn{3}{l}{\textbf{Standard Limb Leads}} \\
Lead I & Left arm & Right arm \\
Lead II & Left leg & Right arm \\
Lead III & Left leg & Left arm \\

\addlinespace
\multicolumn{3}{l}{\textbf{Augmented Limb Leads}} \\
aVR & Right arm & Left arm + left leg \\
aVL & Left arm & Right arm + left leg \\
aVF & Left leg & Left arm + right arm \\

\addlinespace
\multicolumn{3}{l}{\textbf{Precordial Leads}} \\
V$_1$ & Right sternal margin, 4th ICS & Wilson central terminal \\
V$_2$ & Left sternal margin, 4th ICS & Wilson central terminal \\
V$_3$ & Midway between V$_2$ and V$_4$ & Wilson central terminal \\
V$_4$ & Left midclavicular line, 5th ICS & Wilson central terminal \\
V$_5$ & Left anterior axillary line* & Wilson central terminal \\
V$_6$ & Left midaxillary line* & Wilson central terminal \\
V$_7$ & Posterior axillary line* & Wilson central terminal \\
V$_8$ & Posterior scapular line* & Wilson central terminal \\
V$_9$ & Left border of spine* & Wilson central terminal \\
\bottomrule
\end{tabular}}
\caption{Standard 12-Lead ECG Electrode Configuration. The table refers to the clinical textbook~\cite{mirvis2001electrocardiography} Table 13-1.}
\label{tab:ecg_leads}
\end{table}

\begin{table*}[!t]
\centering
\setlength\tabcolsep{0.0055\linewidth}
\renewcommand\arraystretch{1}
 \scalebox{0.84}{\begin{tabular}{crrrrrrrrrrrrrrrrrr}
\toprule
\multirow{2}{*}{\textbf{Method}} & \multicolumn{3}{c}{\textbf{PTBXL-Sub}} & \multicolumn{3}{c}{\textbf{PTBXL-Super}} & \multicolumn{3}{c}{\textbf{PTBXL-Form}} & \multicolumn{3}{c}{\textbf{PTBXL-Rhythm}} & \multicolumn{3}{c}{\textbf{CPSC2018}} & \multicolumn{3}{c}{\textbf{CSN}} \\
                       & \multicolumn{1}{c}{1\%}       & \multicolumn{1}{c}{10\%}     & \multicolumn{1}{c}{100\%}    & \multicolumn{1}{c}{1\%}       & \multicolumn{1}{c}{10\%}     & \multicolumn{1}{c}{100\%}  &  \multicolumn{1}{c}{1\%}       & \multicolumn{1}{c}{10\%}     & \multicolumn{1}{c}{100\%}    &  \multicolumn{1}{c}{1\%}       & \multicolumn{1}{c}{10\%}     & \multicolumn{1}{c}{100\%}   &  \multicolumn{1}{c}{1\%}       & \multicolumn{1}{c}{10\%}     & \multicolumn{1}{c}{100\%}  &  \multicolumn{1}{c}{1\%}       & \multicolumn{1}{c}{10\%}     & \multicolumn{1}{c}{100\%} \\ \midrule
{N/A} &  73.86 & 80.11 & 89.3 & 78.59 & 85.43 & 88.68 & 61.00 & 69.96 & 80.34 & 79.24 & 86.72 & 93.66   &  64.82  &  78.10   &  89.59   &  62.88  &  76.33 &  86.75  \\ 
{Weighted Average} &  73.20 & 80.14 & 89.5 & 79.10 & 86.10 & 89.96 & 60.32 & 70.52 & 81.47 & 78.34 & 86.93 &  93.81  & 64.51  &  80.01 & 90.01  &  66.73   &    80.02                      &         88.11\\ 
{Single-level Grouping}  & 74.14 & 81.10 & 90.40 & 78.87 & 85.70 & 90.17 & 60.74 & 72.67 & 81.77 & 78.95  & 87.25 & 93.00  & 65.49  & 80.63  & 90.64  &   68.70  &   80.21                       &          89.35 \\ \midrule
CLEAR-HUG    & 76.66     & 84.59    & 91.44    & 79.61    & 86.85    & 90.24    & 61.01     & 75.19     & 82.86    & 79.48    & 90.57    & {94.08}   &     66.97    &    82.59     &     91.17     &    72.09   &    82.14    &    89.93\\ \bottomrule
\end{tabular}}
\caption{Comparison of different designs to integrate 12 leads. HUG shows the best performance across 6 datasets.}
\label{tab:res1}
\end{table*}

\begin{table*}[]
\centering
\setlength\tabcolsep{0.0244\linewidth}
\scalebox{0.84}{%
\begin{tabular}{l|lllll}
\toprule
Leads                                             & PTBXL-Sub & PTBXL-Super & PTBXL-Form & PTBXL-Rhythm & \textit{avg.}   \\ \hline
I                                                 & 80.55     & 82.36       & 60.55      & 86.52        & 77.50 \\
I II                                              & 81.27     & 83.53       & 71.86      & 89.88        & 81.64 \\
I II V2                                           & \underline{83.66}     & \underline{84.69}       & \underline{74.42}      & \underline{90.05}        & \underline{83.21} \\
I, II, III, aVR, aVL, aVF                         & 82.78     & 82.69       & 68.03      & 90.14        & 77.83 \\
I, II, III, aVR, aVL, aVF, V1, V2, V3, V4, V5, V6 & \textbf{84.59}     & \textbf{86.85}       & \textbf{75.19}      & \textbf{90.57}        & \textbf{84.30} \\ 
\bottomrule
\end{tabular}%
}
\caption{The performance of missing leads finetuning with the CLEAR pretrained model. We select different lead combinations for downstream tasks. Notably, CLEAR-HUG surpasses previous SOTA methods in most cases with only 2 leads (I, II), compared to Table 1.}
\label{tab:missingleads}
\end{table*}

\subsection{B. Implementation detail}
We present the detailed settings for CLEAR-HUG, as shown in Tab.~\ref{tab:paraclear} and Tab.~\ref{tab:parahug}. During the pretraining stage, all models were trained on 6 A100 GPUs, utilizing 96 cores and 1500GB of memory. For the fine-tuning stage, the experiments were conducted on a single A100 GPU, with 8 cores and 250GB of memory.

\begin{table}[!h]
\centering
\setlength\tabcolsep{0.135\linewidth}
{%
\scalebox{0.85}{\begin{tabular}{c|c}
\toprule
\textbf{Hyperparameters} & \textbf{Values} \\ \midrule
Transformer encoder layers & 12 \\ 
Transformer decoder layers & 4 \\ 
Hidden size & 768 \\ 
MLP size & 1024 \\ 
Attention head number & 8 \\ \midrule
Batch size & 1536 \\
Peak learning rate & 5e-4 \\ 
Minimal learning rate & 1e-5 \\ 
Learning rate scheduler & Cosine \\
Optimizer & AdamW \\
Adam $\beta$ & (0.9, 0.99) \\
Weight decay & 1e-4 \\
Total epochs & 100 \\
Warmup epochs & 10 \\ \bottomrule
\end{tabular}
}}
\caption{Hyperparameters CLEAR Pre-training.}
\label{tab:paraclear}
\end{table}

\begin{table}[!h]
\centering
\setlength\tabcolsep{0.16\linewidth}
{%
\scalebox{0.85}{\begin{tabular}{c|c}
\toprule
\textbf{Hyperparameters} & \textbf{Values} \\ \midrule
Batch size & 256 \\ 
Peak learning rate & 5e-3 \\ 
Minimal learning rate & 1e-5 \\ 
Learning rate scheduler & Cosine \\ 
Optimizer & AdamW \\
Weight decay & 0.05 \\
Layer decay & 0.9 \\
Total epochs & 100 \\
Warmup epochs & 10 \\ \bottomrule
\end{tabular}}}
\caption{Hyperparameters for downstream fine-tuning.}
\label{tab:parahug}
\end{table}

\subsection{C. Ablation of Masking Ratio}
We conducted experiments with varying masking ratios (0.65, 0.8, and 0.95) and evaluated the performance using the average AUC across six datasets: PTBXL-Sub, PTBXL-Supe, PTBXL-Form, and PTBXL-Rhythm. The average AUCs are 84.19, 84.30, and 83.71. A high masking ratio of 80\% achieves optimal results. Notably, the performance across those masking ratios exhibits only marginal differences. We attribute this observation to the sparse attention mechanism, which restricts the model’s focus to a limited subset of tokens during masked token reconstruction, thereby diminishing the impact of higher masking ratios.

\begin{figure*}[!h]
	\centering
	\includegraphics[width=1.0\linewidth]{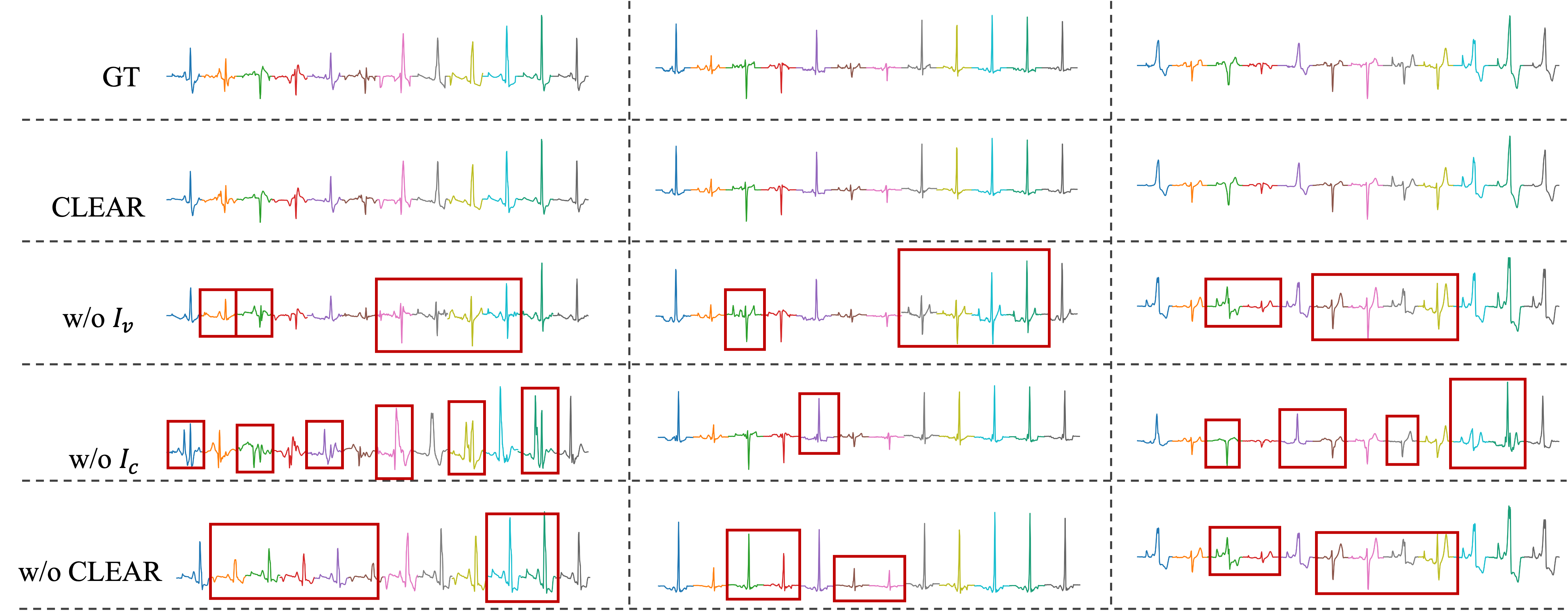
    }
    \caption{{Illustration of proposed CLEAR-HUG framework.}}
    \label{fig:more}
    \vskip -0.1in
\end{figure*}

\subsection{D. Hierarchical Lead Integration}
\label{hierarchical}
We provide more results of the evaluation on different integration strategies for grouping 12 leads as shown in Tab.~\ref{tab:res1}. Our design HUG shows the best performance across 6 datasets. Especially for PTBXL-Sub, CPSC2018, and CSN, our method has a higher improvement in fine-grained classification. The experimental results further prove that our HUG design is effective.



\subsection{E. Missing Leads}
Based on the exploration of missing leads from previous work~\cite{jin2025reading, oh2022lead}, we examine the impact of missing leads on performance and identify failure cases. Results presented in Tab.~\ref{tab:missingleads}. All experiments are conducted in 10\% training data on PTBXL-Sub, PTBXL-Super, PTBXL-Form, and PTBXL-Rhythm datasets. In most cases, downstream performance improves significantly as the number of leads increases, particularly in the Form subsets for disease diagnosis. Notably, our pretrained model demonstrates exceptional performance, surpassing previous SOTA results with only two leads, indicating its strong adaptation to scenarios involving missing leads. Furthermore, we observe performance degradation when leads V1-V6 are missing, which aligns with clinical principles, as certain diseases are best observed from these views. When both bipolar limb leads and precordial leads are included, performance demonstrates improved robustness across all tasks.

\subsection{F. Training Schedule}
\begin{figure}[]
	\centering
	\includegraphics[width=1.0\linewidth]{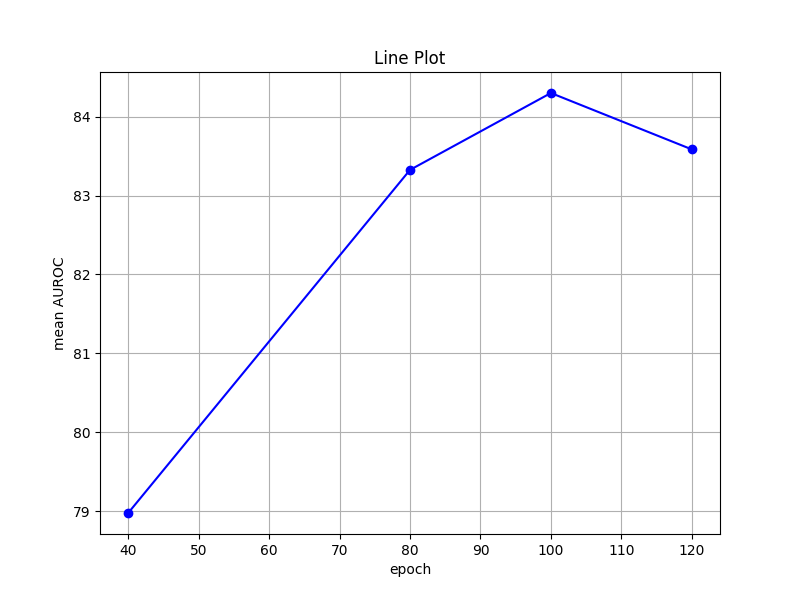}
    \caption{Training schedules. We did experiments on CLEAR pretraining and HUG finetuning under different pretraining epochs. The model under 100 epochs shows the best performance.}
    \label{fig:ckpt}
\end{figure}

Current ablation studies utilize 100-epoch pretraining. Figure~\ref{fig:ckpt} demonstrates that mean AUC scores for PTBXL-Form, PTBXL-Sub, PTBXL-Super, and PTBXL-Rhythm metrics exhibit steady improvement with increasing training duration, achieving peak performance at 100 epochs.

\subsection{G. Efficiency Discussion}

As shown in Tab.~\ref{tab:efficiency_comparison_transposed}, we compare CLEAR-HUG with state-of-the-art eSSL methods such as HeartLang~\cite{jin2025reading} and ST-MEM~\cite{na2024guiding}. We compared the three metrics, flops, namely learnable parameters and GPU memory usage. The result shows that with comparable 48.35M parameters, lower GFLOPs (3.703), and less GPU memory usage(173.71MB), our proposed CLEAR-HUG achieves better performance than both state-of-the-art methods (HeartLang and ST-MEM).

\begin{table}[!h]
\centering

\setlength\tabcolsep{0.038\linewidth}

\scalebox{0.85}{\begin{tabular}{lcccc}
\toprule
 & \textbf{CLEAR-HUG} & \textbf{CLEAR-HUG w/o mask} & \textbf{HeartLang} & \textbf{ST-MEM} \\
\midrule
Parameters(M) & 48.35 & 48.35 & 39.1 & 88.6 \\
FLOPs (G) & 3.703 & 3.732 & 9.796 & 3.767 \\
Memory (MB) & 173.71 & 347.51 & 294.42 & 377.13 \\
Time(ms) & 223 & 278 & 267 & 271 \\

\bottomrule
\end{tabular}}
\caption{Efficiency comparison of CLEAR-HUG and state-of-the-art methods.}
\label{tab:efficiency_comparison_transposed}
\end{table}

\subsection{H. More Visualization of $I_v$ and $I_c$}
We present more visualization of reconstructed signals as shown in Fig.~\ref{fig:more}. CLEAR shows stable reconstruction results.




\end{document}